\documentclass[letterpaper]{article} 
\usepackage{aaai23}  
\usepackage{times}  
\usepackage{helvet}  
\usepackage{courier}  
\usepackage[hyphens]{url}  
\usepackage{graphicx} 
\usepackage{natbib}  
\usepackage{caption} 
\usepackage{algorithm}
\usepackage{algorithmic}

\usepackage{newfloat}
\usepackage{listings}
\DeclareCaptionStyle{ruled}{labelfont=normalfont,labelsep=colon,strut=off} 
\floatstyle{ruled}
\newfloat{listing}{tb}{lst}{}
\floatname{listing}{Listing}
\usepackage{algorithm}
\usepackage{algorithmic}

\usepackage{newfloat}
\usepackage{listings}
\usepackage{booktabs}
\usepackage{multirow}

\usepackage{graphicx}
\usepackage{subfigure}
\usepackage{amsmath,amsthm,amssymb,amsfonts}

\usepackage{color}

\usepackage{colortbl}
\definecolor{LightCyan}{rgb}{0.95,0.9,1}

\DeclareCaptionStyle{ruled}{labelfont=normalfont,labelsep=colon,strut=off} 
\floatstyle{ruled}
\newfloat{listing}{tb}{lst}{}
\floatname{listing}{Listing}
\title{Transformation-Equivariant 3D Object Detection\\ for Autonomous Driving}
\author{
    Hai Wu \textsuperscript{\rm 1,2}\thanks{The work done during internship at Inceptio Technology.}, 
    Chenglu Wen \textsuperscript{\rm 1}\footnote{ Corresponding author.},
    Wei Li \textsuperscript{\rm 2},
    Xin Li \textsuperscript{\rm 3},
    Ruigang Yang \textsuperscript{\rm 2},
    Cheng Wang \textsuperscript{\rm 1}
}
\affiliations{
    \textsuperscript{\rm 1} Xiamen University, 
    \textsuperscript{\rm 2} Inceptio Technology,
    \textsuperscript{\rm 3} Texas A \& M University\\
    wuhai@stu.xmu.edu.cn, \{clwen, cwang\}@xmu.edu.cn, liweimcc@gmail.com,\\ ruigang.yang@inceptio.ai, xinli@tamu.edu
}

\usepackage{bibentry}

\begin{document}

\maketitle

\begin{abstract}
3D object detection received increasing attention in autonomous driving recently. Objects in 3D scenes are distributed with diverse orientations. 
Ordinary detectors do not explicitly model the variations of rotation and reflection transformations. 
Consequently, large networks and extensive data augmentation are required for robust detection.
Recent equivariant networks explicitly model the transformation variations by applying shared networks on multiple transformed point clouds, showing great potential in object geometry modeling. 
However, it is difficult to apply such networks to 3D object detection in autonomous driving due to its large computation cost and slow reasoning speed. 
In this work, we present TED, an efficient \textbf{T}ransformation-\textbf{E}quivariant 3D \textbf{D}etector to overcome the computation cost and speed issues. 
TED first applies a sparse convolution backbone to extract multi-channel transformation-equivariant voxel features; and then aligns and aggregates these equivariant features into lightweight and compact representations for high-performance 3D object detection.
On the highly competitive KITTI 3D car detection leaderboard,
TED \textbf{ranked 1st among all submissions}\footnote{On the date of AAAI deadline, August 15, 2022.} with competitive efficiency. 
\end{abstract}

\section{Introduction}
\label{introduction}
3D object detection plays a crucial role in scene perception of safe autonomous driving~\cite{MVP}.
In recent years, a large number of 3D object detectors have been proposed~\cite{Voxel-RCNN,BtcDet,SFD}.
Most of these methods construct detection frameworks based on ordinary voxels~\cite{SECOND,Voxelnet} or point-based operations~\cite{STD,3DSSD}.

It is desired that the predictions from a 3D detector are equivariant to transformations such as rotations and reflections. 
In other words, when an object changes its orientation in the input points, the detected bounding box of this object should have a same shape but change its orientation accordingly. 
However, most voxel- and point-based approaches do not explicitly model such a transformation equivariance, and could produce unreliable detection results when processing transformed point clouds. 
Some detectors achieve approximate transformation equivariance through data augmentations~\cite{PointRCNN,PV-RCNN}. Their performance, however, heavily relies on generating comprehensive training samples and adopting more complex networks with larger capacity. 
Some other approaches~\cite{TTA} employ a Test Time Augmentation (TTA) scheme, which improves the detection robustness under transformations by running detectors multiple times with transformed inputs.  
Recently, equivariant networks~\cite{EON} have been developed to explicitly model the transformation equivariance. 
They transform input data with multiple rotation bins and model the equivariance by shared convolutional networks.
Both TTA and equivariant design demonstrated promising results. However, both of them require massive computation to process differently transformed point clouds. Consequently, they cannot achieve real-time performance in 3D detection. 
For example, when applying four transformations to PointRCNN~\cite{PointRCNN}, 
the detection time increases from around $0.1$s to more than $0.4$s per detection. 
This severely hinders the application of equivariant 3D detectors to real-time systems such as autonomous driving. 

\begin{figure*}
  \centering
  \includegraphics[width=0.75\textwidth]{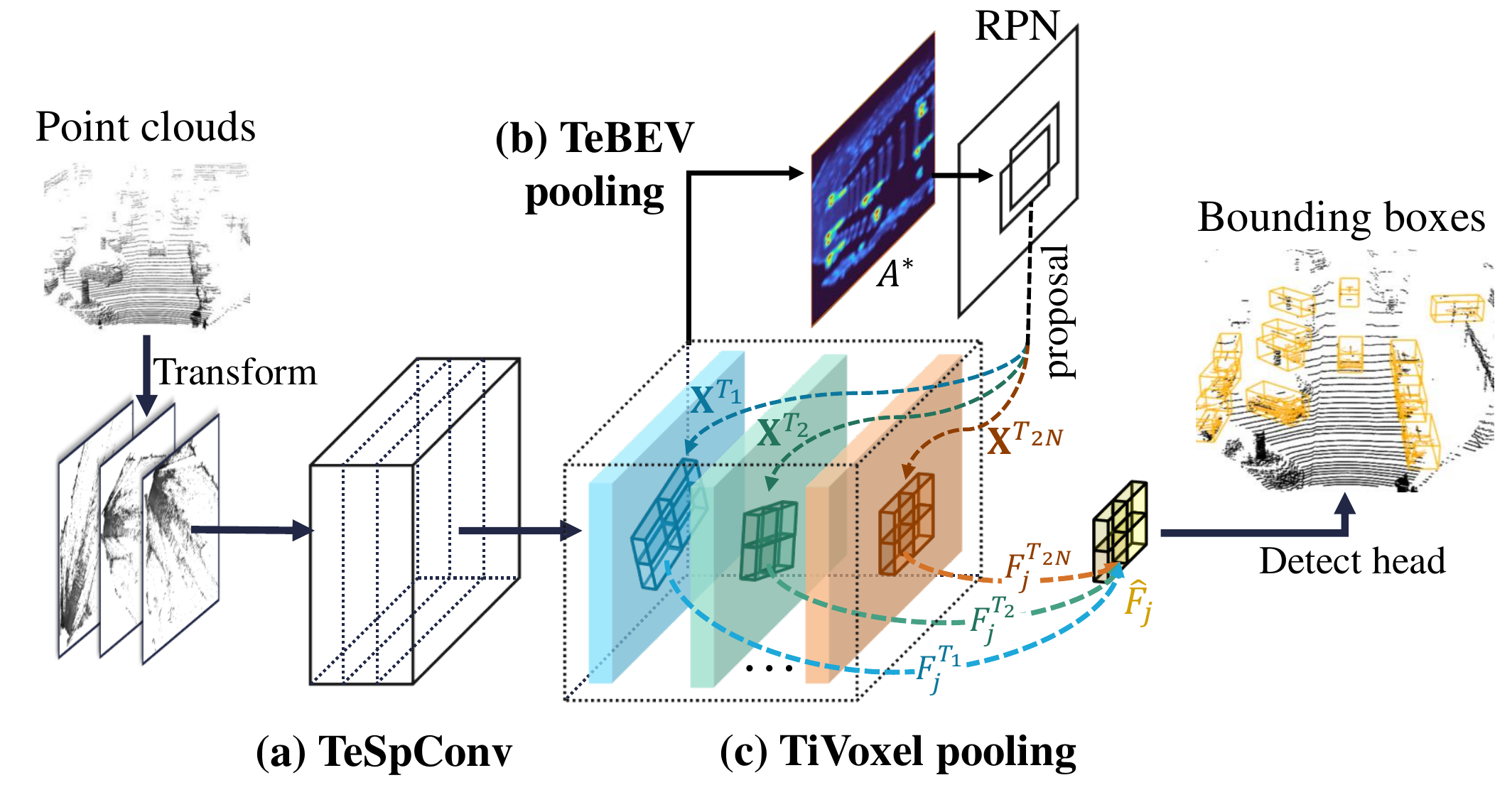}
  \caption{TED framework. (1) The TeSpConv applies shared sparse convolution on multiple transformed point clouds to record the multi-channel transformation-equivariant voxel features. (2) TeBEV pooling aligns and aggregates the scene-level features for proposal generation. (3)
TiVoxel pooling aligns and aggregates the instance-level features for proposal refinement.}
  \label{framework}
\end{figure*}

In this work, we present TED, a \textbf{T}ransformation-\textbf{E}quivariant 3D \textbf{D}etector to tackle this efficiency issue.  
TED first applies a sparse convolution backbone to extract multi-channel transformation-equivariant voxel features. Then TED aligns and aggregates the equivariant features into a lightweight and compact representation for high-performance 3D object detection.
As shown in Fig.~\ref{framework}, TED has three key parts:  (1) the Transformation-equivariant Sparse Convolution (TeSpConv) backbone; (2) Transformation-equivariant Bird Eye View (TeBEV) pooling; and (3) Transformation-invariant Voxel (TiVoxel) pooling. 
TeSpConv applies shared weights on multiple transformed point clouds to record the transformation-equivariant voxel features. TeBEV pooling aligns and aggregates the scene-level equivariant features into lightweight representations for proposal generation.
TiVoxel pooling aligns and aggregates the instance-level invariant features into compact representations for proposal refinement.
In addition, we designed a Distance-Aware data Augmentation (DA-Aug) to enhance geometric knowledge of sparse objects. 
DA-Aug creates sparse training samples from nearby dense objects. 

We conducted comprehensive experiments on both KITTI~\cite{KITTI} and Waymo dataset~\cite{Waymo}. On the KITTI~\cite{KITTI} 3D detection leaderboard, TED attains a performance of \textbf{85.28\% AP} in the moderate Car class with competitive speed, and currently \textbf{ ranks 1st among all the submissions}.  Our technical contributions are summarized as follows: (1) We introduce novel TeBEV pooling and TiVoxel pooling modules that efficiently learn equivariant features from point clouds, leading to better 3D object detection. (2) We propose a new DA-Aug mechanism to create more sparse training samples from nearby dense objects. DA-Aug be used as a general off-the-shelf augmentation method to enhance distant object detection.  
(3) TED sets a new state-of-the-art on both KITTI and Waymo datasets, showing the effectiveness of our design. 

\section{Related Work}
\label{gen_inst}
\paragraph{3D object detection.}
Previous methods convert point clouds into 2D multi-view images to perform 3D object detection~\cite{MV3D,Birdnet}.
Large amounts of recent methods adopt a voxel or point-based detection framework.
By using voxel-based sparse convolution, SECOND~\cite{SECOND}, PointPillars~\cite{PointPillars}, SA-SSD~\cite{SA-SSD}, and SE-SSD~\cite{SE-SSD} perform single-stage 3D object detection, while Voxel-RCNN~\cite{Voxel-RCNN} and SFD~\cite{SFD} perform two-stage detection.
By using point-based set abstraction, 3DSSD~\cite{3DSSD}, SASA~\cite{SASA} and IA-SSD~\cite{IA-SSD} perform single-stage detection, while PointRCNN~\cite{PointRCNN} and STD~\cite{STD} perform two-stage detection.  PV-RCNN~\cite{PV-RCNN} and CT3D~\cite{CT3D} generate and refine object proposal using both voxel-based and point-based operations. Some recent methods generate pseudo points~\cite{SFD} or virtual points~\cite{MVP} from RGB images for voxel-based multi-modal 3D object detection. 
We employ the voxel-based two-stage detection pipeline in our design. The difference is that we extend the pipeline with a transformation-equivariant design, leading to higher detection accuracy.

\paragraph{Transformation equivariance modeling.}
Recently, a great variety of transformation (translation, rotation and reflection) equivariant networks has been presented~\cite{SCNN,GECNNs,GECN,3DSCNNs}. 
Some point-based SO(3)~\cite{SO3,EPN} and voxel-based SE(3) equivariant~\cite{3DSCNNs} networks are designed to process 3D data. Some equivariant networks are designed for object detection~\cite{ReDet,EON} and pose estimation ~\cite{SSC}. The most similar work is the EON~\cite{EON} designed for indoor scenes; however, it did not consider the efficiency. Our TED is mostly designed for real-time object detection from outdoor scene.

\paragraph{Object augmentation for 3D detection.}
The ``copy-and-paste'' based augmentation (GT-aug) is widely used in recent works~\cite{PointRCNN, PV-RCNN, Voxel-RCNN}. LiDAR-Aug~\cite{LiDAR-Aug} addressed the occlusion problem in GT-aug. By swapping  objects' parts, PA-Aug~\cite{PA-Aug} and SE-SSD~\cite{SE-SSD} create diverse objects for data augmentation. Different from them, we enhance distant object detection by creating sparse training samples from nearby dense objects.

\section{Preliminaries}
\label{preli}
\paragraph{Transformation equivariance and invariance.}
Given an operation $f: X\rightarrow Y$ and a transformation group $G$, the equivariance is defined as:
\begin{equation}
    f[T_g^X(x)] = T_g^Y[f(x)], \forall x\in X, \forall g\in G,
\label{eq1}
\end{equation}
where $T_g^X$ and $T_g^Y$ refer to a transformation action in the $X$ and $Y$ space, respectively. When $T_g^Y$ is an identity matrix, the equivariance becomes invariance.
This paper investigates 3D object detection in the autonomous driving scenario, in which the transformation mostly occurs on a road plane. Without loss of generality, we consider the transformation on the 2D BEV plane. Formally, we consider the transformation group $G$ as the semidirect product of the 2D translation group $(\mathbb{R}^2, +)$ and the 2D rotation reflection group $K$ as $G \cong (\mathbb{R}^2, +)\rtimes K$. The rotation reflection group $K$ consists of a reflection group $(\{\pm 1\},*)$ and a discrete rotation group $O_N$. Group $K$ contains $N$ discrete rotations by angles multiple of angle resolution $\beta$ in the case of reflection; therefore, $K$ is a discrete subgroup of order $2N$.
For an input cloud $P$, this paper tries to find a detector $D^\theta(\cdot)$ with parameters $\theta$ to detect bounding boxes $B$, satisfying the transformation equivariance as:
\begin{equation}
    D^\theta [T_g(P)] = T_g[D^\theta (P)],
\label{eq2}
\end{equation}
where $T_g$ is the transformation action in group $G$.

\paragraph{3D object detection.} 
Given an input point cloud $P=\{p_i\}_i$, 3D object detection aims to find all objects represented by 3D BBs $B=\{b_i\}_i$. Each box $b_i$ is encoded by coordinates, size and orientation. Recent works takes into LiDAR point clouds~\cite{PointGNN,PointRCNN,PV-RCNN} or image data~\cite{F-ConvNet,F-PointNet,CLOCs,SFD,MVP}.
For a broader comparison, in this paper, we construct a single-modal TED-S and a multi-modal TED-M detector from the State-of-the-art Voxel-RCNN~\cite{Voxel-RCNN}. The difference between TED-S and TED-M is that the TED-S takes into only LiDAR points, while TED-M takes into both LiDAR points and RGB image pseudo points generated by depth estimation algorithm~\cite{PENet}.
We follow the region-based detection framework~\cite{Voxel-RCNN}. It consists of a sparse convolution backbone, a 2D Region Proposal Network (RPN) and a proposal refinement branch.

\section{Our Method}
Lots of 3D detectors build upon voxel~\cite{SECOND,Voxelnet,Voxel-RCNN} or point-based~\cite{PointRCNN,STD,3DSSD,SASA} operations.
The regular voxel and point-based operations are not rotation and reflection-equivariant.
Recent work improves the transformation robustness by explicitly modeling rotation equivariance or applying Test Time Augmentation (TTA). Nevertheless, both of them require massive computation to process differently transformed point clouds, and cannot achieve real-time performance in 3D detection.
To tackle this, we present TED, which is both transformation-equivariant and efficient. This is done by a simple yet effective design: let TeSpConv stack multi-channel transformation-equivariant voxel features, while TeBEV pooling and TiVoxel pooling align and aggregate the equivariant features into lightweight scene-level and instance-level representations for efficient and effective 3D object detection.

\subsection{Transformation-Equivariant Voxel Backbone}
To efficiently encode raw points into transformation-equivariant features, we first design a Transformation-equivariant Sparse Convolution (TeSpConv) backbone. The TeSpConv is constructed from the widely used sparse convolution (SpConv)~\cite{SpConv,Voxel-RCNN}. Similar to CNNs, SpConv is translation-equivariant.
However, the SpConv is not equivariant to rotation and reflection. 
To address this, we extend SpConv to rotation and reflection equivariant by adding transformation channels. Similar to a 2D counterpart~\cite{LSCNNs,ReDet,TI-POOLING}, the equivariance is achieved by two aspects: (1) weights highly shared between transformation channels; (2) transformation of input points with different rotation angles and reflection. 
Formally, based on $2N$ transformation actions $\{T_i\}_{i=1}^{2N}\subset K$, we transform point clouds $P$ into $2N$ sets of different points $\{P^{T_i}\}_{i=1}^{2N}$. After that, all the point sets are divided into voxel sets $\{\hat{P}^{T_i}\}_{i=1}^{2N}$, respectively. In each voxel, the raw features are calculated as the mean of point-wise features of all inside points. 
We use shared SpConv $\varphi(\cdot)$ to encode the voxels $\{\hat{P}^{T_i}\}_{i=1}^{2N}$ into transformation-equivariant voxel features $\{V^{T_i}\}_{i=1}^{2N}$ as
\begin{equation}
    V^{T_i} = \varphi(\hat{P}^{T_i}), i=1,2,...,2N.
\label{eq4}
\end{equation}
Note that, for the multi-modal setting, we encode the features of pseudo-points by the same network architecture. 
Compared with the voxel features encoded by regular sparse convolution, features $\{V^{T_i}\}_{i=1}^{2N}$ contain diverse features under different rotation and reflection transformations.

\subsection{Transformation-Equivariant BEV Pooling}
\begin{figure}
  \centering
  \includegraphics[width=0.95\columnwidth]{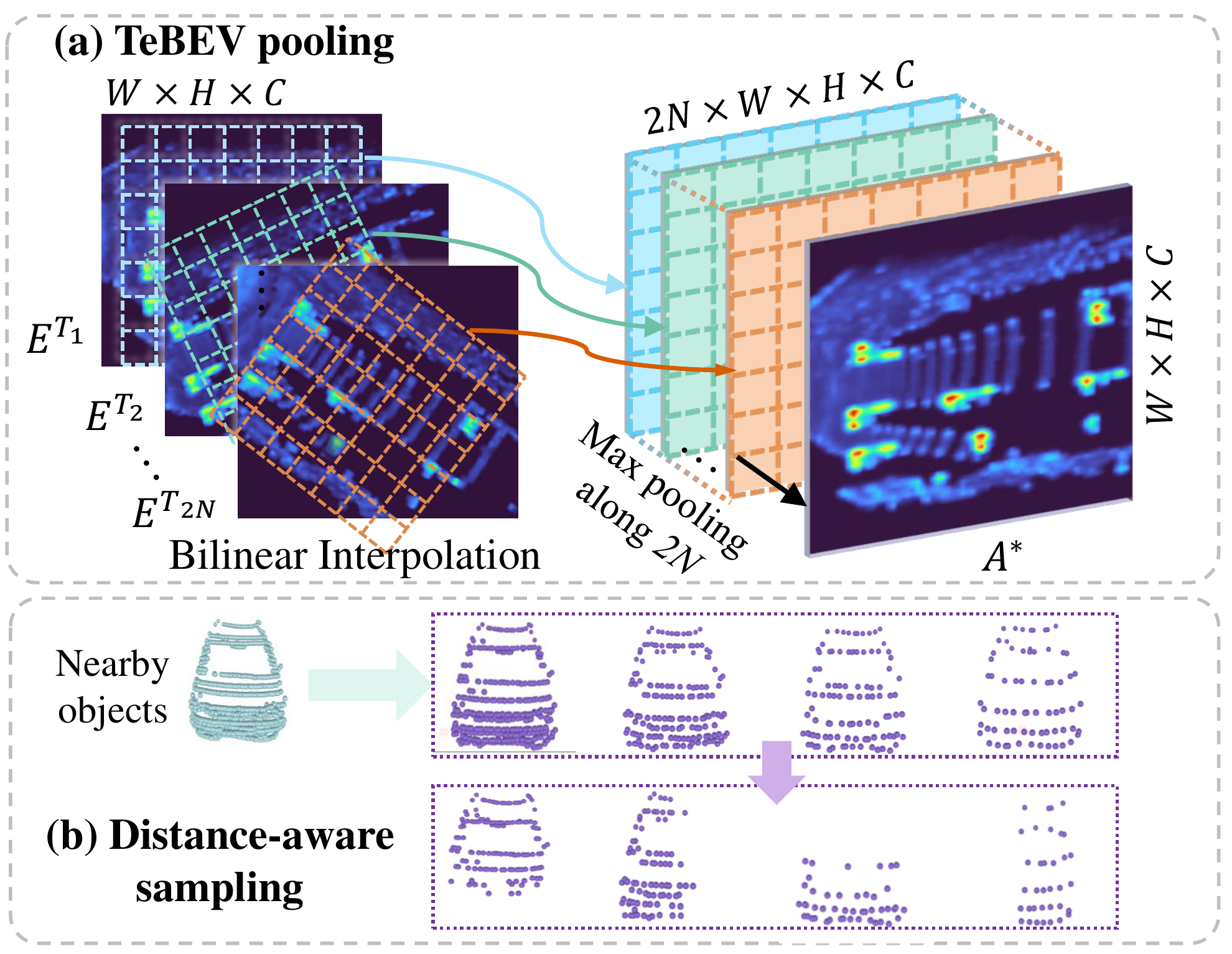}
  \caption{(a) TeBEV pooling: aligning and aggregating the scene-level transformation-equivariant voxel features by bilinear interpolation and max pooling. (b) DA-Aug: creating distant sparse training samples from nearby dense objects by simulating LiDAR scanning and occlusion.}
  \label{bevalign}
\end{figure}

The voxel features $\{V^{T_i}\}_{i=1}^{2N}$ contain a large number of transformation channels; thus, directly feeding them into RPN introduces huge additional computation and requires larger GPU memory. To address this, we propose the TeBEV pooling, which aligns and aggregates the scene-level voxel features into a compact BEV map by bilinear interpolation and max-pooling (see Fig.~\ref{bevalign} (a)). 
Specifically, the voxel features $\{V^{T_i}\}_{i=1}^{2N}$ are first compressed into BEV features $\{E^{T_i}\}_{i=1}^{2N}$ along height dimension. Since the BEV features are obtained under different transformations, it is necessary to align them into the same coordinate system. We first generate a set of scene-level grid points $X^{T_1}$ in the coordinate system $E^{T_1}$. Based on the transformation actions $ \{T_i\}_{i=1}^{2N}$, we transform the grid points into the BEV coordinate system to generate a set of new grid points $\{X^{T_i}\}_{i=1}^{2N}$. After that, we apply a series of bilinear interpolation $\mathcal{I}(\cdot,\cdot)$ on the BEV maps to obtain a set of aligned features $\{A^{T_i}\}_{i=1}^{2N}$ as:
\begin{equation}
    A^{T_i} = \mathcal{I}(X^{T_i},E^{T_i}), i=1,2,...,2N.
\label{eq5}
\end{equation}
If the boundary pixel in $E^{T_1}$ has no corresponding pixel in $E^{T_2},...,E^{T_N}$, the interpolated result will be padded by zero.
For efficiency, we apply a max-pooling $\mathcal{M}(\cdot)$ on the $2N$ aligned feature maps to obtain a compact representation $A^*$ as 
\begin{equation}
    A^* = \mathcal{M}(A^{T_1},A^{T_2},...,A^{T_{2N}}).
\label{eq6}
\end{equation}
The lightweight feature $A^*$ is put into RPN to efficiently generate a set of object proposals $B^*$. 

\subsection{Transformation-Invariant Voxel Pooling}

Lots of recent detectors apply Region of Interest (RoI) pooling operation that extracts instance-level transformation-invariant features from scene-level transformation-equivariant backbone features for proposal refinement~\cite{EON,ReDet}.
Nevertheless, directly applying such a pooling operation to extract features from our backbone is infeasible. The reasons are: (1) the proposal $B^*$ in the coordinate system $T^1$ is unaligned with the voxel features $\{V^{T_i}\}_{i=1}^{2N}$ transformed by different $T^i$. (2) The voxel features from our TeSpConv contains multiple transformation channels, and directly feeding the extracted features to the detection head needs huge additional computation and GPU memory.
To address these issues, we propose TiVoxel pooling, which aligns and aggregates the instance-level voxel features into a compact feature vector by multi-grid pooling and cross-grid attention.

\paragraph{Multi-grid pooling.} 
The proposal $B^*$ is obtained in the coordinate system $T^1$, while the 
backbone features are obtained under different transformations. Hence a proposal alignment is needed before pooling. We first generate a set of local grid points using proposal $B^*$ in the coordinate system $T^1$, then transform the grid points into the coordinate system of every channel in $\{A^{T_i}\}_{i=1}^{2N}$ based on the transformation actions $\{T^2,...,T^{2N}\}$, and finally extract transformation-invariant local features by the $2N$ sets of transformed grid points.
Specifically, For a proposal in $B^*$, based on the transformation actions $ \{T_i\}_{i=1}^{2N}$, we first generate $2N$ sets of instance-level grid points $\{\mathbf{X}^{T_i}\}_{i=1}^{2N}$, where $\mathbf{X}^{T_i} = \{X^{T_i}_j\}_{j=1}^{J}\subset \mathbb{R}^3$. $J$ denotes the grid points number in each set. By using the $2N$ sets of grid points, we extract multiple instance-level features $\{\mathbf{F}^{T_i}\}_{i=1}^{2N}$ from $\{V^{T_i}\}_{i=1}^{2N}$ as:
\begin{equation}
    \mathbf{F}^{T_i} = VSA(\mathbf{X}^{T_i}, V^{T_i}), i=1,2,...,2N,
\label{eq7}
\end{equation}
where $VSA$ refers the Voxel Set Abstraction~\cite{Voxel-RCNN} module, and $\mathbf{F}^{T_i}=\{F^{T_i}_j\}_{j=1}^{J}\subset \mathbb{R}^{1\times C}$. $C$ denotes the number of grid-wise feature channels.

\begin{table*}
	\centering

    \resizebox{\textwidth}{!}{
	\begin{tabular}{l| l | l l l  | l l l |  l l l  }
		\hline
		\multirow{2}*{Method}         &\multirow{2}*{Modality}  &\multicolumn{3}{c|}{Car 3D (R40)}   &\multicolumn{3}{c|}{Pedestrian 3D (R40)}   &\multicolumn{3}{c}{Cyclist 3D (R40)}   \\ 
		                               &                         &Easy &Mod. &Hard     &Easy &Mod. &Hard       &Easy &Mod. &Hard              \\
		\hline 
        PV-RCNN\cite{PV-RCNN}             &LiDAR   &92.57 &84.83 &82.69       &64.26 &56.67 &51.91         &88.88 &71.95 &66.78 \\
        Voxel-RCNN\cite{Voxel-RCNN}       &LiDAR   &92.38 &85.29 &82.86       &-     &-     &-             &-     &-     &-  \\
        CT3D\cite{CT3D}                  &LiDAR    &92.85 &85.82 &83.46       &65.73 &58.56 &53.04         &91.99 &71.60 &67.34       \\
        SE-SSD\cite{SE-SSD}               &LiDAR   &\textbf{93.19} &86.12 &83.31        &-     &-     &-            &-     &-     &-       \\	   
        BtcDet\cite{BtcDet}               &LiDAR   &93.15 &86.28 &83.86       &69.39 &61.19 &55.86         &91.45 &74.70 &70.08 \\

        \rowcolor{LightCyan} TED-S (Ours)                     &LiDAR   & 93.05 &\textbf{87.91} &\textbf{85.81} &\textbf{72.38} &\textbf{67.81} &\textbf{63.54}  &\textbf{93.09} &\textbf{75.77} &\textbf{71.20}  \\
		\hline
        F-PointNet\cite{F-PointNet}       &LiDAR+RGB   &83.76 &70.92 &63.65        &70.00 &61.32 &53.59        &77.15 &56.49 &53.37   \\
        F-ConvNet\cite{F-ConvNet}         &LiDAR+RGB   &89.02 &78.80 &77.09        &-     &-     &-            &-     &-     &-   \\
        CLOCs\cite{CLOCs}                 &LiDAR+RGB   &92.78 &85.94 &83.25        &-     &-     &-            &-     &-     &-       \\
        EPNet++\cite{EPNet++}             &LiDAR+RGB   &92.51 &83.17 &82.27        &73.77 &65.42 &59.13        &86.23 &63.82 &60.02       \\
        SFD\cite{SFD}                     &LiDAR+RGB   &\textbf{95.52} &88.27 &85.57        &72.94 &66.69 &61.59        &93.39 &72.95 &67.26       \\
        \rowcolor{LightCyan} TED-M (Ours)                     &LiDAR+RGB   &95.25 &\textbf{88.94} &\textbf{86.73} &\textbf{74.73} &\textbf{69.07} &\textbf{63.63}  &\textbf{95.20} &\textbf{76.17} &\textbf{71.59}  \\
		\hline 
	\end{tabular}
    }
    \caption{3D detection results on the KITTI val set, the best LiDAR-only and multi-modal methods are in bold respectively. Our TED-S outperforms all of the previous LiDAR-only methods, while TED-M outperforms all of the previous multi-modal methods.}
    \label{table:kittival}
\end{table*}

\paragraph{Cross-grid attention.}
Our transformation-invariant $\{\mathbf{F}^{T_i}\}_{i=1}^{2N}$ contains multiple instance-level features. To encode better local geometry, we apply a cross-grid attention operation to further aggregate the multiple features into a single more compact transformation-invariant feature vector. In detail, for $j$th grid point, we concatenate $2N$ grid-wise features $F_j = concat(F_j^{T_1},...,F_j^{T_{2N}})$, where $F_j\in \mathbb{R}^{2N\times C}$. Then we have $\mathbf{Q}_j =  F_j \mathbf{W}^q$, $\mathbf{K}_j =  F_j\mathbf{W}^k$, $\mathbf{V}_j =  F_j\mathbf{W}^v$, where $\mathbf{W}^q,\mathbf{W}^k,\mathbf{W}^v \in  \mathbb{R}^{C\times C}$ are linear projections. The grid-wise feature is calculated by 
\begin{equation}
    \hat{F}_j = softmax(\frac{\mathbf{Q}_j(\mathbf{K}_j)^T}{\sqrt{C}} )\mathbf{V}_j.
\end{equation}
We average the feature along $2N$, and obtain features $\hat{\mathbf{F}}=\{\hat{F}_j\}_{j=1}^{J}\subset \mathbb{R}^{1\times C}$. These features are then flattened into a single feature vector to perform object proposal refinement similar to recent detectors~\cite{PV-RCNN,Voxel-RCNN}.

\subsection{Distance-Aware Data Augmentation}
\label{da-aug}
The geometry incompleteness of distant objects commonly results in a huge detection performance drop.  
To address this, we increase the geometric knowledge of distant sparse objects by creating sparse training samples from nearby dense objects. 
A simple method is to apply random sampling or farthest point sampling (FPS). However, it destroys the distribution pattern of the point clouds scanned by LiDAR. 

To address this, we propose a distance-aware sampling strategy, which considers the scanning mechanics of LiDAR and scene occlusion.
Specifically, given a nearby ground truth box with position $C^g$ and inside points $\{P^g_i\}_i$, we add a random distance offset $\Delta\alpha$ as $C^g:=C^g+\Delta\alpha$, $P^g_i:=P^g_i+\Delta\alpha$. Then we convert $\{P^g_i\}_i$ into a spherical coordinate system and voxelize them into spherical voxels based on the angular resolution of LiDAR. In each voxel, the point nearest to the voxel center is retained as a sampled point. After that, we obtain a set of sampled points, which has a similar distribution pattern as real scanned points (as shown in Fig.~\ref{bevalign} (b)). Since incomplete data caused by occlusion is common in real scenes, we also randomly remove some parts to simulate the occlusion. During training, similar to GT-AUG~\cite{PointRCNN}, we add the sampled points and bounding box into the training sample for data augmentation.

\section{Experiments}
 
\label{exp}
\subsection{Dataset and Metrics}
\label{dataset}
\textbf{(1) KITTI Dataset.}
The KITTI~\cite{KITTI} Dataset contains 7481 training and 7518 testing frames. We follow recent work~\cite{Voxel-RCNN, SFD} to divide the training data into a train split of 3712 frames and a val split of 3769 frames. We report the results using the 3D Average Precision (AP) (\%) under 40 recall thresholds (R40). 
\textbf{(2) Waymo Open Dataset (WOD).}
WOD~\cite{Waymo} contains 798 training and 202 validation sequences. The official metrics are mean Average Precision (mAP (L1), mAP (L2)) with heading (mAPH (L1) and mAPH (L2)), where L1 and L2 denote the difficulty level.

\begin{table*}
	\centering

    \resizebox{\textwidth}{!}{
	\begin{tabular}{l | l | l l l  | l l l |  l l l  }
		\hline
		\multirow{2}*{Method}         &\multirow{2}*{Modality}  &\multicolumn{3}{c|}{Car 3D (R40)}   &\multicolumn{3}{c|}{Pedestrian 3D (R40)}   &\multicolumn{3}{c}{Cyclist 3D (R40)}   \\
		                               &                         &Easy &Mod. &Hard     &Easy &Mod. &Hard       &Easy &Mod. &Hard              \\
		\hline
        PV-RCNN\cite{PV-RCNN}             &LiDAR   &90.25 &81.43 &76.82        &52.17 &43.29 &40.29        &78.60 &63.71 &57.65 \\
        Voxel-RCNN\cite{Voxel-RCNN}       &LiDAR  &90.90 &81.62 &77.06        &-     &-     &-            &-     &-     &-       \\
        PV-RCNN++\cite{PV-RCNN2}          &LiDAR  &90.14 &81.88 &77.15       &54.29 &47.19 &43.49         &82.22 &67.33 &60.04 \\
        SE-SSD\cite{SE-SSD}               &LiDAR  &91.49 &82.54 &77.15        &-     &-     &-            &-     &-     &-       \\	   
        BtcDet\cite{BtcDet}               &LiDAR   &90.64 &82.86 &78.09        &47.80 &41.63 &39.30        &82.81 &68.68 &61.81 \\
		\hline
        F-ConvNet\cite{F-ConvNet}         &LiDAR+RGB   &87.36 &76.39 &66.69        &52.16 &43.38 &38.8         &81.98 &65.07 &56.54   \\
        CLOCs\cite{CLOCs}                 &LiDAR+RGB   &88.94 &80.67 &77.15        &-     &-     &-            &-     &-     &-       \\
        EPNet++\cite{EPNet++}             &LiDAR+RGB   &91.37 &81.96 &76.71        &52.79 &44.38 &41.29        &76.15 &59.71 &53.67       \\
        SFD\cite{SFD}                     &LiDAR+RGB   &\textbf{91.73} &84.76 &77.92        &-     &-     &-            &-     &-     &-       \\

        \rowcolor{LightCyan} TED-M (Ours)                     &LiDAR+RGB   & 91.61 &\textbf{85.28} &\textbf{80.68} &\textbf{55.85} &\textbf{49.21} &\textbf{46.52} &\textbf{88.82} &\textbf{74.12}   &\textbf{66.84} \\
		\hline
	\end{tabular}
    }
    \caption{3D detection results on the KITTI test set, the best methods are in bold. TED-M outperforms all of the previous methods in all moderate Car, Pedestrian and Cyclist classes. }
    \label{table:kittitest}
\end{table*}

\begin{table*}
	\centering

    \resizebox{\textwidth}{!}{
	\begin{tabular}{l|c c| c c | c c| c c | c c| c c }
		\hline
		\multirow{2}*{Methods}           &\multicolumn{2}{c|}{Vehicle(L1)} &\multicolumn{2}{c|}{Vehicle(L2)} &\multicolumn{2}{c|}{Ped.(L1)} &\multicolumn{2}{c|}{Ped.(L2)} &\multicolumn{2}{c|}{Cyc.(L1)} &\multicolumn{2}{c}{Cyc.(L2)}\\
                                         &mAP     &mAPH     &mAP     &mAPH     &mAP     &mAPH     &mAP     &mAPH     &mAP     &mAPH     &mAP   &mAPH\\
        \hline               
	    $\ddagger$Voxel-RCNN~\cite{Voxel-RCNN}      &77.43   &76.71    &68.73   &68.24    &76.37   &68.21    &67.92   &60.40    &68.74   &67.56    &66.46   &65.35 \\
	    $\dagger$PV-RCNN\cite{PV-RCNN}   &77.51   &76.89    &68.98   &68.41    &75.01   &65.65    &66.04   &57.61   & 67.81   &66.35    &65.39   &63.98 \\
	    BtcDet\cite{BtcDet}              &78.58   &78.06    &70.10   &69.61    &-       &-        &-       &-        &-       &-        &-       &-     \\
	    PV-RCNN++\cite{PV-RCNN2}         &78.79   &78.21    &70.26   &69.71    &76.67   &67.15    &68.51   &59.72    &68.98   &67.63    &66.48   &65.17 \\
	    PDA\cite{PDA}                    &-       &-        &-       &69.98     &-       &-        &-       &60.00      &-      &-         &-      &67.88\\
        \rowcolor{LightCyan}TED-S (ours)       &\textbf{79.26}   &\textbf{78.73}    &\textbf{70.50}   &\textbf{70.07}    &\textbf{82.62}   &\textbf{76.66}    &\textbf{73.50}   &\textbf{68.03}    &\textbf{74.11}   &\textbf{72.94}    &\textbf{71.46}   &\textbf{70.32} \\
		\hline
	\end{tabular}
    }
    \caption{3D detection results on the Waymo validation set (using two returns and a single frame). $\dagger$ : reported in ~\cite{PV-RCNN2}.$\ddagger$ : reproduced results by us using their open-source code. The best results are in bold. }	
    \label{table:waymoval}
\end{table*}

\subsection{Setup Details }
\label{setup}
\paragraph{Network setup.}
To demonstrate the universality and superiority of our TED, we construct a LiDAR-only TED-S and a multi-modal TED-M (see Section Preliminaries). We adopted 3 different rotation transformations and 2 different reflection transformations along the x-axis. For each detector on each dataset, we train a single model to detect three classes of objects.

\paragraph{Training and inference setup.}
For the KITTI dataset, we adopt the detection range, proposal number and NMS threshold as the same as the baseline detector Voxel-RCNN~\cite{Voxel-RCNN}.
For the Waymo dataset, we adopted the same setting as PV-RCNN++\cite{PV-RCNN2}.
We train all the detectors on two 3090 GPU cards with a batch size of four and an Adam optimizer with a learning rate of 0.01.
For data augmentation, without rotation and reflection data augmentation, our method can achieve a high detection performance. With the data augmentation, we obtain slightly better results. Scaling, local augmentation and ground-truth sampling are also used. To increase more geometric knowledge of distant sparse objects, we apply our proposed DA-Aug in the ground-truth sampling.

\subsection{Comparison with State-of-the-art}
\paragraph{KITTI validation set.}
We first conducted experiments on the KITTI validation set. The results are shown in Table~\ref{table:kittival}. For the most important metric, moderate Car AP(40), our TED outperforms previous methods under both LiDAR-only and multi-modal settings by a large margin. Specifically, our LiDAR-only TED-S and multi-modal TED-M improve baseline Voxel-RCNN~\cite{Voxel-RCNN} by 2.62\% and 3.65\% respectively, and outperforms the SOTA LiDAR-only BtcDet~\cite{BtcDet} and multi-modal SFD~\cite{SFD} by 1.63\% and 0.67\% respectively. It's worth noting that our TED-S and TED-M attain high performance in Pedestrian and Cyclist classes.  Specifically,  our TED-S outperforms the SOTA LiDAR-only BtcDet~\cite{BtcDet} by 6.62\% and 1.07\% in the  Pedestrian and Cyclist classes respectively. Our TED-M also outperforms the SOTA multi-modal SFD~\cite{BtcDet} by 2.38\% and 3.22\% on  Pedestrian and Cyclist respectively. The performance gains are mostly derived from the transformation-equivariant design that learns better geometric features of objects, leading to better detection performance. 

\begin{table}

    \small
	\centering
	\begin{tabular}{c | c    c  | c  c }
		\hline
		\multirow{2}*{Rotation number}  &\multicolumn{2}{c|}{TED-S}   &\multicolumn{2}{c}{TED-M} \\  
		 &Car 3D AP  & FPS  &Car 3D AP & FPS\\
		\hline
		1     &85.73                &20.1        &86.64              &18.3\\
        2     &87.13                &14.4        &87.82              &13.3\\
        3     &87.91                &11.1        &\textbf{88.94}              &10.6\\
        4     &\textbf{87.95}                &9.3         &88.89              &8.3\\
		\hline
	\end{tabular}
    \caption{Results on the KITTI val set by using different rotation numbers. }

    \label{table:rotnum}
\end{table}

\paragraph{KITTI test set.}
To further demonstrate the advance of our design, we further submitted the test set results produced by our most advanced TED-M to the KITTI online server (We followed the KITTI regulation: only submit the best results to the online test server). As shown in Table~\ref{table:kittitest}, our TED-M surpass the baseline Voxel-RCNN~\cite{Voxel-RCNN} by 3.66\% on the moderate Car class. Besides, TED-M outperforms all of the previous methods in all of the moderate Car, Pedestrian and Cyclist classes. 
As of August 15, 2022, on the highly competitive KITTI 3D detection benchmark, our TED-M ranks 1st in Car class. These results fully demonstrated the effectiveness of our method.

\paragraph{Waymo validation set.}
For the waymo dataset, we follow Voxel-RCNN~\cite{Voxel-RCNN} and PV-RCNN~\cite{PV-RCNN}, only using a single frame. We trained our LiDAR-only TED-S on the Waymo dataset using first and second LiDAR returns. The results are summarized in Table~\ref{table:waymoval}. Our TED-S improves the baseline Voxel-RCNN~\cite{Voxel-RCNN} by 1.83\%, 7.63\% and 4.97\% on Vehicle(L2), Pedestrian(L2) and Cyclist(L2) metrics respectively. TED-S also outperforms all previous methods on all metrics. The results further demonstrate the effectiveness of our design.

\begin{table}

    \small
	\centering
    \resizebox{\columnwidth}{!}{
	\begin{tabular}{c | c | c c c  |  c }
		\hline
		\multirow{2}*{Method}   &\multirow{2}*{Backbone}    & \multirow{2}*{DA-Aug}  &TeBEV       &TiVoxel    &\multirow{2}*{TED-S}  \\
	     &  & & pooling &pooling & \\
		\hline
        Baseline &SpConv      &          &           &               &85.29        \\
        Ours     &SpConv      &\checkmark&           &               &85.72        \\
        Ours     &TeSpConv    &\checkmark&\checkmark &               &86.33       \\
        Ours     &TeSpConv    &\checkmark&\checkmark &\checkmark     &\textbf{87.91}        \\
		\hline
	\end{tabular}
    }
    \caption{Results on the KITTI val set by applying different designed components. }

    \label{table:comp}
\end{table}

\subsection{Ablation Study}
\label{ablation}
To examine the effectiveness of each component in our method and choose the best hyper-parameter, we conduct a series of ablation experiments on the Car class of the KITTI validation set. 

\noindent\textbf{Rotation number.}
In our method, the number of rotation transformations $N$ is a hyper-parameter. We choose the best $N$ by considering both the detection accuracy and efficiency. The results using different $N$ are shown in Table~\ref{table:rotnum}. We observe that the detection performance of the $N=3$ is close to $N=4$. For computational efficiency, we adopt $N=3$. It can achieve real-time performance with a running speed of around 11 FPS (single 3090 GPU). Compared with other popular algorithms (see Table ~\ref{table:timecom}), TED attains competitive efficiency under both LiDAR-only and multi-modal settings.

\noindent\textbf{Component analysis.}
Next, we verify each component in our design under both LiDAR-only and multi-modal settings. The LiDAR-only baseline is the Voxel-RCNN~\cite{Voxel-RCNN}. We also constructed a multi-modal baseline: multi-modal Voxel-RCNN, which takes into both LiDAR point clouds and RGB image pseudo points. As shown in Table~\ref{table:comp}, by adding our proposed components, both the LiDAR-only and multi-modal baselines are improved by a large margin. Specifically, by adding DA-Aug, TeBEV pooling and TiVoxel pooling, the LiDAR-only baseline is improved by 0.43\%, 0.61\% and 1.58\%, respectively, and the multi-modal baseline is also improved by 0.49\%, 0.62\% and 1.7\%, respectively. This highlights the effectiveness of our designed components.

\begin{figure} 
\centering
  \includegraphics[width=\columnwidth]{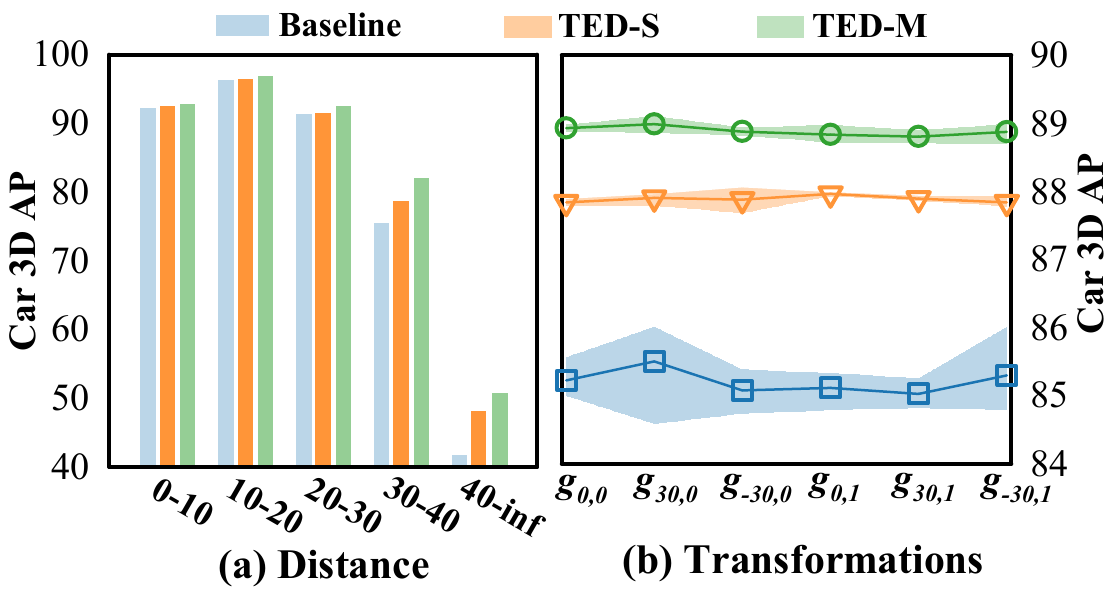}
\caption{ (a) Performance at different detection distances to the ego-vehicle. (b) Performance and error bar by applying different transformations to the input point cloud. }
\label{dis-rot}
\end{figure}

\noindent\textbf{Performance breakdown.}
To investigate in what cases our model improves the baseline most, we evaluate the detection performance based on the different distances. The results are shown in Fig.~\ref{dis-rot} (a). Our LiDAR-only TED-S has a significant improvement for faraway objects.  
This improvement comes from the fine-grained geometric features generated from our TED. Our multi-modal TED-M has a further improvement thanks to the appearance features from the RGB image.

\noindent\textbf{Robustness to transformation.}
Our design is more robust to transformation variations.
To demonstrate such property, we applied six different transformations $\{g_{i,j}\}_{i,j}$ to the input point clouds, where $i$ denotes rotation angle along z-axis and $j$ denotes whether flip points along x-axis. We trained our TED-S, TED-M and baseline model multiple times and report the validation results with an error bar in Fig.~\ref{dis-rot} (b). Our method shows more stable predictions under different transformations.
The reason is that our TED-S and TED-M explicitly model the rotation and reflection variations by TeSpConv, TeBEV pooling and TiVoxel pooling, thus having more robust performance. 

\begin{table}

    \small
	\centering

	\begin{tabular}{l| c| c| c  }
		\hline
		Method    & AOS@Easy  &AOS@Moderate  &AOS@Hard  \\
		\hline 
        Baseline  &97.77      &94.75         &92.13     \\
        TED-S     &98.41      &96.10         &93.61     \\
        TED-M    &\textbf{99.21}      &\textbf{97.10}         &\textbf{94.83}     \\
		\hline
		
	\end{tabular}
    \caption{Orientation accuracy on the KITTI val set. }
    \label{table:orien}
\end{table}

\noindent\textbf{Orientation accuracy.}
Thanks to the rotation-equivariant property, our model can estimate more accurate object orientation. We conduct experiments on the KITTI validation set using the official average orientation similarity (AOS). 
As shown in Table~\ref{table:orien}, our TED-S and TED-M show significant AOS improvements compared with the baseline model, demonstrating the effectiveness of our method.

\noindent\textbf{Comparison with TTA.}
We conducted an experiment to compare our method with TTA~\cite{TTA}, which improves 3D object detection by applying multiple transformations to inputs and merging the results by weighted box fusion. The results are shown in Table~\ref{table:TTAvsOurs}. TTA boosts the baseline by about 1\%~AP climbing with 0.27s inference speed. Thanks to our efficient and effective TeBEV pooling and TiVoxel pooling design, our method gets better performance with 2.2\%~AP improved and only 0.09s consumed for inference.

\noindent\textbf{Comparison with previous equivariant detector.}
We also conducted an experiment to compare our method with previous equivariant detector EON~\cite{EON}. The results are shown in Table~\ref{table:EON}. Our method not only obtains better detection precision, but also runs much faster than the EON. The reason is that our detector is built upon more efficient sparse convolution, TeBEV pooling and TiVoxel pooling modules. The results demonstrate that our design is much applicable to autonomous driving.

\begin{table}

    \small
	\centering

	\begin{tabular}{l | c | c  }
		\hline
		Method    & Modiality  &FPS \\
		\hline 
		PV-RCNN\cite{PV-RCNN}  &LiDAR-only      &10.25  \\
        Voxel-RCNN\cite{Voxel-RCNN}  &LiDAR-only      &25  \\

        CasA\cite{CasA}  &LiDAR-only      &12.5 \\
        TED-S  &LiDAR-only      &11.1  \\
        \hline
        F-PointNet\cite{F-PointNet} &LiDAR+RGB      &5.9     \\
        EPNet\cite{EPNet}  &LiDAR+RGB      &10     \\
        SFD\cite{SFD}  &LiDAR+RGB      &10.2     \\
        TED-M  &LiDAR+RGB     &10.6    \\
		\hline
	\end{tabular}
    \caption{Compared with recent two-stage methods, TED-S and TED-M obtain competitive efficiency under LiDAR-only and multi-modal settings respectively. }
    \label{table:timecom}
\end{table}

\begin{table}[htbp] 
    \small
	\centering
	\begin{tabular}{l |c | c}
		\hline
		 Method (w DA-Aug) & AP 3D (Car mod.)  &  Time (s)\\
		\hline
         VoxelRCNN                               &85.72           &\textbf{0.04}\\
         VoxelRCNN + TTA     &86.65        &0.27\\
         TED-S (Ours)         &\textbf{87.91} &0.09\\
		\hline
	\end{tabular}
	\caption{Moderate car 3D detection results on the KITTI val set by using TTA and our method.}
	\label{table:TTAvsOurs}
\end{table}

\begin{table}[htbp] 
    \small
	\centering
	\begin{tabular}{l |c c c| c}
		\hline
		 Method & Car &Pedestrian &Cyclist & Time (s)\\
		\hline
         EON      &78.61 &61.08 &73.36 &0.4*\\
         TED-S (Ours)       &\textbf{87.91} &\textbf{67.81} &\textbf{75.77} &\textbf{0.09}\\
		\hline
	\end{tabular}
	\caption{Comparison with previous rotation-equivariant detector EON~\cite{EON}. * denotes that time is measured by our-self.}
	\label{table:EON}
\end{table}

\section{Conclusion}
\label{conclusion}
This paper proposed TED, a high-performance 3D object detector for point clouds.
TED encodes transformation-equivariant voxel features into compact scene-level and instance-level representations for object proposal generation and refinement.
This design is efficient and learns the geometric features of objects better. On the competitive KITTI 3D object detection benchmark, TED ranks 1st among all submissions, demonstrating its effectiveness.

\noindent\textbf{Limitations.} 
(1) Our TED design is not strictly transformation-equivariant due to the discretization of transformation and voxelization of input. By using more transformations and smaller voxels, TED will be closer to fully equivariant. But this comes with a higher computational cost. 
(2) Considering that incorporating more transformations increases computational cost, we did not include scaling transformations which are observed much less in practical scenes. 
(3) Compared with the Voxel-RCNN baseline, our method requires about 2$\times$ overall GPU memory.

\noindent\textbf{Acknowledgement.}
This work was supported in part by the National Natural Science Foundation of China (No.62171393), and the Fundamental Research Funds for the Central Universities (No.20720220064).


{\small
\bibliography{aaai23}
}
\end{document}